# Active Learning with Statistical Models

**David A. Cohn**   COHN@HARLEQUIN.COM
**Zoubin Ghahramani**   ZOUBIN@CS.TORONTO.EDU
**Michael I. Jordan**   JORDAN@PSYCHE.MIT.EDU
*Center for Biological and Computational Learning*
*Dept. of Brain and Cognitive Sciences*
*Massachusetts Institute of Technology*
*Cambridge, MA 02139 USA*

## Abstract

For many types of machine learning algorithms, one can compute the statistically "optimal" way to select training data. In this paper, we review how optimal data selection techniques have been used with feedforward neural networks. We then show how the same principles may be used to select data for two alternative, statistically-based learning architectures: mixtures of Gaussians and locally weighted regression. While the techniques for neural networks are computationally expensive and approximate, the techniques for mixtures of Gaussians and locally weighted regression are both efficient and accurate. Empirically, we observe that the optimality criterion sharply decreases the number of training examples the learner needs in order to achieve good performance.

## 1. Introduction

The goal of machine learning is to create systems that can improve their performance at some task as they acquire experience or data. In many natural learning tasks, this experience or data is gained interactively, by taking actions, making queries, or doing experiments. Most machine learning research, however, treats the learner as a passive recipient of data to be processed. This "passive" approach ignores the fact that, in many situations, the learner's most powerful tool is its ability to act, to gather data, and to influence the world it is trying to understand. Active learning is the study of how to use this ability effectively.

Formally, active learning studies the closed-loop phenomenon of a learner selecting actions or making queries that influence what data are added to its training set. Examples include selecting joint angles or torques to learn the kinematics or dynamics of a robot arm, selecting locations for sensor measurements to identify and locate buried hazardous wastes, or querying a human expert to classify an unknown word in a natural language understanding problem.

When actions/queries are selected properly, the data requirements for some problems decrease drastically, and some NP-complete learning problems become polynomial in computation time (Angluin, 1988; Baum & Lang, 1991). In practice, active learning offers its greatest rewards in situations where data are expensive or difficult to obtain, or when the environment is complex or dangerous. In industrial settings each training point may take days to gather and cost thousands of dollars; a method for optimally selecting these points could offer enormous savings in time and money.

There are a number of different goals which one may wish to achieve using active learning. One is optimization, where the learner performs experiments to find a set of inputs that maximize some response variable. An example of the optimization problem would be finding the operating parameters that maximize the output of a steel mill or candy factory. There is an extensive literature on optimization, examining both cases where the learner has some prior knowledge of the parameterized functional form and cases where the learner has no such knowledge; the latter case is generally of greater interest to machine learning practitioners. The favored technique for this kind of optimization is usually a form of response surface methodology (Box & Draper, 1987), which performs experiments that guide hill-climbing through the input space.

A related problem exists in the field of adaptive control, where one must learn a control policy by taking actions. In control problems, one faces the complication that the value of a specific action may not be known until many time steps after it is taken. Also, in control (as in optimization), one is usually concerned with the performing well *during* the learning task and must trade of exploitation of the current policy for exploration which may improve it. The subfield of dual control (Fe'ldbaum, 1965) is specifically concerned with finding an optimal balance of exploration and control while learning.

In this paper, we will restrict ourselves to examining the problem of supervised learning: based on a set of potentially noisy training examples $\mathcal{D} = \{(x_i, y_i)\}_{i=1}^{m}$, where $x_i \in X$ and $y_i \in Y$, we wish to learn a general mapping $X \to Y$. In robot control, the mapping may be $state \times action \to new\ state$; in hazard location it may be $sensor\ reading \to target\ position$. In contrast to the goals of optimization and control, the goal of supervised learning is to be able to efficiently and accurately predict $y$ for a given $x$.

In active learning situations, the learner itself is responsible for acquiring the training set. Here, we assume it can iteratively select a new input $\tilde{x}$ (possibly from a constrained set), observe the resulting output $\tilde{y}$, and incorporate the new example $(\tilde{x}, \tilde{y})$ into its training set. This contrasts with related work by Plutowski and White (1993), which is concerned with filtering an existing data set. In our case, $\tilde{x}$ may be thought of as a query, experiment, or action, depending on the research field and problem domain. The question we will be concerned with is how to choose which $\tilde{x}$ to try next.

There are many heuristics for choosing $\tilde{x}$, including choosing places where we don't have data (Whitehead, 1991), where we perform poorly (Linden & Weber, 1993), where we have low confidence (Thrun & Möller, 1992), where we expect it to change our model (Cohn, Atlas, & Ladner, 1990, 1994), and where we previously found data that resulted in learning (Schmidhuber & Storck, 1993). In this paper we will consider how one may select $\tilde{x}$ in a statistically "optimal" manner for some classes of machine learning algorithms. We first briefly review how the statistical approach can be applied to neural networks, as described in earlier work (MacKay, 1992; Cohn, 1994). Then, in Sections 3 and 4 we consider two alternative, statistically-based learning architectures: mixtures of Gaussians and locally weighted regression. Section 5 presents the empirical results of applying statistically-based active learning to these architectures. While optimal data selection for a neural network is computationally expensive and approximate, we find that optimal data selection for the two statistical models is efficient and accurate.

## 2. Active Learning – A Statistical Approach

We begin by defining $P(x, y)$ to be the unknown joint distribution over $x$ and $y$, and $P(x)$ to be the known marginal distribution of $x$ (commonly called the *input distribution*). We denote the learner's output on input $x$, given training set $\mathcal{D}$ as $\hat{y}(x; \mathcal{D})$.[1] We can then write the expected error of the learner as follows:

$$\int_x E_T \left[ (\hat{y}(x; \mathcal{D}) - y(x))^2 \,|\, x \right] P(x) dx, \tag{1}$$

where $E_T[\cdot]$ denotes expectation over $P(y|x)$ and over training sets $\mathcal{D}$. The expectation inside the integral may be decomposed as follows (Geman, Bienenstock, & Doursat, 1992):

$$\begin{aligned} E_T \left[ (\hat{y}(x; \mathcal{D}) - y(x))^2 \,|\, x \right] &= E \left[ (y(x) - E[y|x])^2 \right] \\ &\quad + (E_{\mathcal{D}}[\hat{y}(x; \mathcal{D})] - E[y|x])^2 \\ &\quad + E_{\mathcal{D}} \left[ (\hat{y}(x; \mathcal{D}) - E_{\mathcal{D}}[\hat{y}(x; \mathcal{D})])^2 \right] \end{aligned} \tag{2}$$

where $E_{\mathcal{D}}[\cdot]$ denotes the expectation over training sets $\mathcal{D}$ and the remaining expectations on the right-hand side are expectations with respect to the conditional density $P(y|x)$. It is important to remember here that in the case of active learning, the distribution of $\mathcal{D}$ may differ substantially from the joint distribution $P(x, y)$.

The first term in Equation 2 is the variance of $y$ given $x$ — it is the *noise* in the distribution, and does not depend on the learner or on the training data. The second term is the learner's *squared bias*, and the third is its *variance*; these last two terms comprise the mean squared error of the learner with respect to the regression function $E[y|x]$. When the second term of Equation 2 is zero, we say that the learner is *unbiased*. We shall assume that the learners considered in this paper are approximately unbiased; that is, that their squared bias is negligible when compared with their overall mean squared error. Thus we focus on algorithms that minimize the learner's error by minimizing its variance:

$$\sigma_{\hat{y}}^2 \equiv \sigma_{\hat{y}}^2(x) = E_{\mathcal{D}} \left[ (\hat{y}(x; \mathcal{D}) - E_{\mathcal{D}}[\hat{y}(x; \mathcal{D})])^2 \right]. \tag{3}$$

(For readability, we will drop the explicit dependence on $x$ and $\mathcal{D}$ — unless denoted otherwise, $\hat{y}$ and $\sigma_{\hat{y}}^2$ are functions of $x$ and $\mathcal{D}$.) In an active learning setting, we will have chosen the $x$-component of our training set $\mathcal{D}$; we indicate this by rewriting Equation 3 as

$$\sigma_{\hat{y}}^2 = \left\langle (\hat{y} - \langle \hat{y} \rangle)^2 \right\rangle,$$

where $\langle \cdot \rangle$ denotes $E_{\mathcal{D}}[\cdot]$ given a fixed $x$-component of $\mathcal{D}$. When a new input $\tilde{x}$ is selected and queried, and the resulting $(\tilde{x}, \tilde{y})$ added to the training set, $\sigma_{\hat{y}}^2$ should change. We will denote the expectation (over values of $\tilde{y}$) of the learner's new variance as

$$\left\langle \tilde{\sigma}_{\hat{y}}^2 \right\rangle = E_{\mathcal{D} \cup (\tilde{x}, \tilde{y})} \left[ \sigma_{\hat{y}}^2 | \tilde{x} \right]. \tag{4}$$

---

1. We present our equations in the univariate setting. All results in the paper apply equally to the multivariate case.

### 2.1 Selecting Data to Minimize Learner Variance

In this paper we consider algorithms for active learning which select data in an attempt to minimize the value of Equation 4, integrated over $X$. Intuitively, the minimization proceeds as follows: we assume that we have an estimate of $\sigma_{\hat{y}}^2$, the variance of the learner at $x$. If, for some new input $\tilde{x}$, we knew the conditional distribution $P(\tilde{y}|\tilde{x})$, we could compute an estimate of the learner's new variance at $x$ given an additional example at $\tilde{x}$. While the true distribution $P(\tilde{y}|\tilde{x})$ is unknown, many learning architectures let us approximate it by giving us estimates of its mean and variance. Using the estimated distribution of $\tilde{y}$, we can estimate $\left\langle \tilde{\sigma}_{\hat{y}}^2 \right\rangle$, the expected variance of the learner after querying at $\tilde{x}$.

Given the estimate of $\left\langle \tilde{\sigma}_{\hat{y}}^2 \right\rangle$, which applies to a given $x$ and a given query $\tilde{x}$, we must integrate $x$ over the input distribution to compute the integrated average variance of the learner. In practice, we will compute a Monte Carlo approximation of this integral, evaluating $\left\langle \tilde{\sigma}_{\hat{y}}^2 \right\rangle$ at a number of *reference points* drawn according to $P(x)$. By querying an $\tilde{x}$ that minimizes the average expected variance over the reference points, we have a solid statistical basis for choosing new examples.

### 2.2 Example: Active Learning with a Neural Network

In this section we review the use of techniques from Optimal Experiment Design (OED) to minimize the estimated variance of a neural network (Fedorov, 1972; MacKay, 1992; Cohn, 1994). We will assume we have been given a learner $\hat{y} = f_{\hat{w}}()$, a training set $\mathcal{D} = \{(x_i, y_i)\}_{i=1}^{m}$ and a parameter vector estimate $\hat{w}$ that maximizes some likelihood measure given $\mathcal{D}$. If, for example, one assumes that the data were produced by a process whose structure matches that of the network, and that noise in the process outputs is normal and independently identically distributed, then the negative log likelihood of $\hat{w}$ given $\mathcal{D}$ is proportional to

$$S^2 = \frac{1}{m} \sum_{i=1}^{m} (y_i - \hat{y}(x_i))^2 .$$

The maximum likelihood estimate for $\hat{w}$ is that which minimizes $S^2$.

The estimated output variance of the network is

$$\sigma_{\hat{y}}^2 \approx S^2 \left( \frac{\partial \hat{y}(x)}{\partial w} \right)^T \left( \frac{\partial^2 S^2}{\partial w^2} \right)^{-1} \left( \frac{\partial \hat{y}(x)}{\partial w} \right), \text{ (MacKay, 1992)}$$

where the true variance is approximated by a second-order Taylor series expansion around $S^2$. This estimate makes the assumption that $\partial \hat{y}/\partial w$ is locally linear. Combined with the assumption that $P(y|x)$ is Gaussian with constant variance for all $x$, one can derive a closed form expression for $\left\langle \tilde{\sigma}_{\hat{y}}^2 \right\rangle$. See Cohn (1994) for details.

In practice, $\partial \hat{y}/\partial w$ may be highly nonlinear, and $P(y|x)$ may be far from Gaussian; in spite of this, empirical results show that it works well on some problems (Cohn, 1994). It has the advantage of being grounded in statistics, and is optimal given the assumptions. Furthermore, the expectation is differentiable with respect to $\tilde{x}$. As such, it is applicable in continuous domains with continuous action spaces, and allows hillclimbing to find the $\tilde{x}$ that minimizes the expected model variance.

For neural networks, however, this approach has many disadvantages. In addition to relying on simplifications and assumptions which hold only approximately, the process is computationally expensive. Computing the variance estimate requires inversion of a $|w| \times |w|$ matrix for each new example, and incorporating new examples into the network requires expensive retraining. Paass and Kindermann (1995) discuss a Markov-chain based sampling approach which addresses some of these problems. In the rest of this paper, we consider two "non-neural" machine learning architectures that are much more amenable to optimal data selection.

## 3. Mixtures of Gaussians

The mixture of Gaussians model is a powerful estimation and prediction technique with roots in the statistics literature (Titterington, Smith, & Makov, 1985); it has, over the last few years, been adopted by researchers in machine learning (Cheeseman et al., 1988; Nowlan, 1991; Specht, 1991; Ghahramani & Jordan, 1994). The model assumes that the data are produced by a mixture of $N$ multivariate Gaussians $g_i$, for $i = 1, ..., N$ (see Figure 1).

In the context of learning from random examples, one begins by producing a joint density estimate over the input/output space $X \times Y$ based on the training set $\mathcal{D}$. The EM algorithm (Dempster, Laird, & Rubin, 1977) can be used to efficiently find a locally optimal fit of the Gaussians to the data. It is then straightforward to compute $\hat{y}$ given $x$ by conditioning the joint distribution on $x$ and taking the expected value.

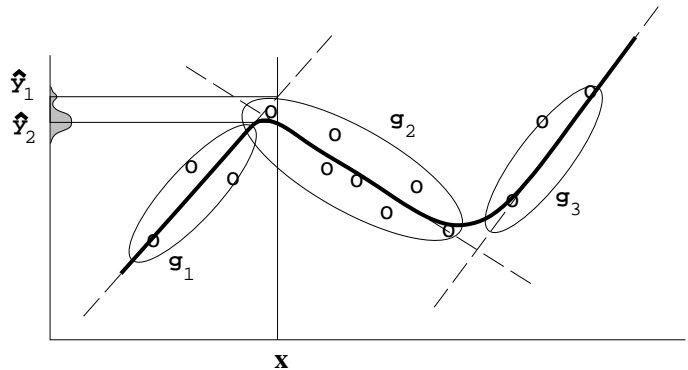

Figure 1: Using a mixture of Gaussians to compute $\hat{y}$. The Gaussians model the data density. Predictions are made by mixing the conditional expectations of each Gaussian given the input $x$.

One benefit of learning with a mixture of Gaussians is that there is no fixed distinction between inputs and outputs — one may specify any subset of the input-output dimensions, and compute expectations on the remaining dimensions. If one has learned a forward model of the dynamics of a robot arm, for example, conditioning on the outputs automatically gives a model of the arm's inverse dynamics. With the mixture model, it is also straightforward to compute the mode of the output, rather than its mean, which obviates many of the problems of learning direct inverse models (Ghahramani & Jordan, 1994).

For each Gaussian $g_i$ we will denote the input/output means as $\mu_{x,i}$ and $\mu_{y,i}$ and variances and covariances as $\sigma^2_{x,i}, \sigma^2_{y,i}$ and $\sigma_{xy,i}$ respectively. We can then express the probability of point $(x,y)$, given $g_i$ as

$$P(x,y|i) = \frac{1}{2\pi\sqrt{|\Sigma_i|}} \exp\left[-\frac{1}{2}(\mathbf{x}-\mu_i)^T \Sigma_i^{-1}(\mathbf{x}-\mu_i)\right] \quad (5)$$

where we have defined

$$\mathbf{x} = \begin{bmatrix} x \\ y \end{bmatrix} \quad \mu_i = \begin{bmatrix} \mu_{x,i} \\ \mu_{y,i} \end{bmatrix} \quad \Sigma_i = \begin{bmatrix} \sigma^2_{x,i} & \sigma_{xy,i} \\ \sigma_{xy,i} & \sigma^2_{y,i} \end{bmatrix}.$$

In practice, the true means and variances will be unknown, but can be estimated from data via the EM algorithm. The (estimated) conditional variance of $y$ given $x$ is then

$$\sigma^2_{y|x,i} = \sigma^2_{y,i} - \frac{\sigma^2_{xy,i}}{\sigma^2_{x,i}}.$$

and the conditional expectation $\hat{y}_i$ and variance $\sigma^2_{\hat{y},i}$ given $x$ are:

$$\hat{y}_i = \mu_{y,i} + \frac{\sigma_{xy,i}}{\sigma^2_{x,i}}(x-\mu_{x,i}), \qquad \sigma^2_{\hat{y},i} = \frac{\sigma^2_{y|x,i}}{n_i}\left(1 + \frac{(x-\mu_{x,i})^2}{\sigma^2_{x,i}}\right). \quad (6)$$

Here, $n_i$ is the amount of "support" for the Gaussian $g_i$ in the training data. It can be computed as

$$n_i = \sum_{j=1}^{m} \frac{P(x_j,y_j|i)}{\sum_{k=1}^{N} P(x_j,y_j|k)}.$$

The expectations and variances in Equation 6 are mixed according to the probability that $g_i$ has of being responsible for $x$, prior to observing $y$:

$$h_i \equiv h_i(x) = \frac{P(x|i)}{\sum_{j=1}^{N} P(x|j)},$$

where

$$P(x|i) = \frac{1}{\sqrt{2\pi\sigma^2_{x,i}}} \exp\left[-\frac{(x-\mu_{x,i})^2}{2\sigma^2_{x,i}}\right]. \quad (7)$$

For input $x$ then, the conditional expectation $\hat{y}$ of the resulting mixture and its variance may be written:

$$\hat{y} = \sum_{i=1}^{N} h_i\,\hat{y}_i, \qquad \sigma^2_{\hat{y}} = \sum_{i=1}^{N} \frac{h_i^2 \sigma^2_{y|x,i}}{n_i}\left(1 + \frac{(x-\mu_{x,i})^2}{\sigma^2_{x,i}}\right),$$

where we have assumed that the $\hat{y}_i$ are independent in calculating $\sigma^2_{\hat{y}}$. Both of these terms can be computed efficiently in closed form. It is also worth noting that $\sigma^2_{\hat{y}}$ is only one of many variance measures we might be interested in. If, for example, our mapping is stochastically multivalued (that is, if the Gaussians overlapped significantly in the $x$ dimension), we may wish our prediction $\hat{y}$ to reflect the most likely $y$ value. In this case, $\hat{y}$ would be the mode, and a preferable measure of uncertainty would be the (unmixed) variance of the individual Gaussians.

### 3.1 Active Learning with a Mixture of Gaussians

In the context of active learning, we are assuming that the input distribution $P(x)$ is known. With a mixture of Gaussians, one interpretation of this assumption is that we know $\mu_{x,i}$ and $\sigma^2_{x,i}$ for each Gaussian. In that case, our application of EM will estimate only $\mu_{y,i}$, $\sigma^2_{y,i}$, and $\sigma_{xy,i}$.

Generally however, knowing the input distribution will not correspond to knowing the actual $\mu_{x,i}$ and $\sigma^2_{x,i}$ for each Gaussian. We may simply know, for example, that $P(x)$ is uniform, or can be approximated by some set of sampled inputs. In such cases, we must use EM to estimate $\mu_{x,i}$ and $\sigma^2_{x,i}$ in addition to the parameters involving $y$. If we simply estimate these values from the training data, though, we will be estimating the joint distribution of $P(\tilde{x}, y|i)$ instead of $P(x, y|i)$. To obtain a proper estimate, we must correct Equation 5 as follows:

$$P(x, y|i) = P(\tilde{x}, y|i) \frac{P(x|i)}{P(\tilde{x}|i)}. \tag{8}$$

Here, $P(\tilde{x}|i)$ is computed by applying Equation 7 given the mean and $x$ variance of the training data, and $P(x|i)$ is computed by applying the same equation using the mean and $x$ variance of a set of reference data drawn according to $P(x)$.

If our goal in active learning is to minimize variance, we should select training examples $\tilde{x}$ to minimize $\left\langle \tilde{\sigma}^2_{\hat{y}} \right\rangle$. With a mixture of Gaussians, we can compute $\left\langle \tilde{\sigma}^2_{\hat{y}} \right\rangle$ efficiently. The model's estimated distribution of $\tilde{y}$ given $\tilde{x}$ is explicit:

$$P(\tilde{y}|\tilde{x}) = \sum_{i=1}^{N} \tilde{h}_i P(\tilde{y}|\tilde{x}, i) = \sum_{i=1}^{N} \tilde{h}_i \mathcal{N}(\hat{y}_i(\tilde{x}), \sigma^2_{y|x,i}(\tilde{x})),$$

where $\tilde{h}_i \equiv h_i(\tilde{x})$, and $\mathcal{N}(\mu, \sigma^2)$ denotes the normal distribution with mean $\mu$ and variance $\sigma^2$. Given this, we can model the change in each $g_i$ separately, calculating its expected variance given a new point sampled from $P(\tilde{y}|\tilde{x}, i)$ and weight this change by $\tilde{h}_i$. The new expectations combine to form the learner's new expected variance

$$\left\langle \tilde{\sigma}^2_{\hat{y}} \right\rangle = \sum_{i=1}^{N} \frac{h_i^2 \left\langle \tilde{\sigma}^2_{y|x,i} \right\rangle}{n_i + \tilde{h}_i} \left( 1 + \frac{(x - \mu_{x,i})^2}{\sigma^2_{x,i}} \right) \tag{9}$$

where the expectation can be computed exactly in closed form:

$$\left\langle \tilde{\sigma}^2_{y,i} \right\rangle = \frac{n_i \sigma^2_{y,i}}{n_i + \tilde{h}_i} + \frac{n_i \tilde{h}_i \left( \sigma^2_{y|\tilde{x},i} + (\hat{y}_i(\tilde{x}) - \mu_{y,i})^2 \right)}{(n_i + \tilde{h}_i)^2}, \quad \left\langle \tilde{\sigma}^2_{y|x,i} \right\rangle = \left\langle \tilde{\sigma}^2_{y,i} \right\rangle - \frac{\left\langle \tilde{\sigma}^2_{xy,i} \right\rangle}{\sigma^2_{x,i}},$$

$$\left\langle \tilde{\sigma}_{xy,i} \right\rangle = \frac{n_i \sigma_{xy,i}}{n_i + \tilde{h}_i} + \frac{n_i \tilde{h}_i (\tilde{x} - \mu_{x,i})(\hat{y}_i(\tilde{x}) - \mu_{y,i})}{(n_i + \tilde{h}_i)^2}, \quad \left\langle \tilde{\sigma}^2_{xy,i} \right\rangle = \left\langle \tilde{\sigma}_{xy,i} \right\rangle^2 + \frac{n_i^2 \tilde{h}_i^2 \sigma^2_{y|\tilde{x},i} (\tilde{x} - \mu_{x,i})^2}{(n_i + \tilde{h}_i)^4}.$$

If, as discussed earlier, we are also estimating $\mu_{x,i}$ and $\sigma^2_{x,i}$, we must take into account the effect of the new example on those estimates, and must replace $\mu_{x,i}$ and $\sigma^2_{x,i}$ in the above equations with

$$\tilde{\mu}_{x,i} = \frac{n_i \mu_{x,i} + \tilde{h}_i \tilde{x}}{n_i + \tilde{h}_i}, \quad \tilde{\sigma}^2_{x,i} = \frac{n_i \sigma^2_{x,i}}{n_i + \tilde{h}_i} + \frac{n_i \tilde{h}_i (\tilde{x} - \mu_{x,i})^2}{(n_i + \tilde{h}_i)^2}.$$

We can use Equation 9 to guide active learning. By evaluating the expected new variance over a reference set given candidate $\tilde{x}$, we can select the $\tilde{x}$ giving the lowest expected model variance. Note that in high-dimensional spaces, it may be necessary to evaluate an excessive number of candidate points to get good coverage of the potential query space. In these cases, it is more efficient to differentiate Equation 9 and hillclimb on $\partial \left\langle \tilde{\sigma}_y^2 \right\rangle / \partial \tilde{x}$ to find a locally maximal $\tilde{x}$. See, for example, (Cohn, 1994).

## 4. Locally Weighted Regression

Model-based methods, such as neural networks and the mixture of Gaussians, use the data to build a parameterized model. After training, the model is used for predictions and the data are generally discarded. In contrast, "memory-based" methods are non-parametric approaches that explicitly retain the training data, and use it each time a prediction needs to be made. Locally weighted regression (LWR) is a memory-based method that performs a regression around a point of interest using only training data that are "local" to that point. One recent study demonstrated that LWR was suitable for real-time control by constructing an LWR-based system that learned a difficult juggling task (Schaal & Atkeson, 1994).

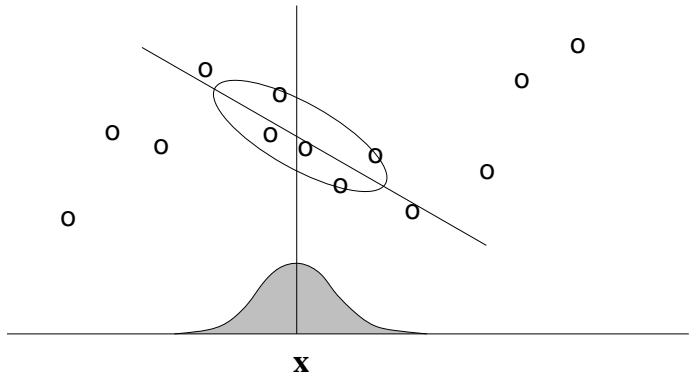

Figure 2: In locally weighted regression, points are weighted by proximity to the current $x$ in question using a kernel. A regression is then computed using the weighted points.

We consider here a form of locally weighted regression that is a variant of the LOESS model (Cleveland, Devlin, & Grosse, 1988). The LOESS model performs a linear regression on points in the data set, weighted by a kernel centered at $x$ (see Figure 2). The kernel shape is a design parameter for which there are many possible choices: the original LOESS model uses a "tricubic" kernel; in our experiments we have used a Gaussian

$$h_i(x) \equiv h(x - x_i) = \exp(-k(x - x_i)^2),$$

where $k$ is a smoothing parameter. In Section 4.1 we will describe several methods for automatically setting $k$.

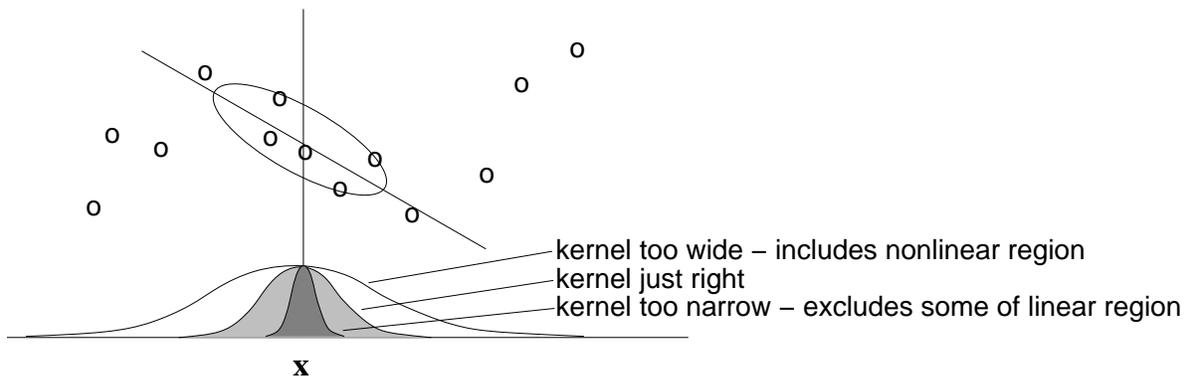

Figure 3: The estimator variance is minimized when the kernel includes as many training points as can be accommodated by the model. Here the linear LOESS model is shown. Too large a kernel includes points that degrade the fit; too small a kernel neglects points that increase confidence in the fit.

For brevity, we will drop the argument $x$ for $h_i(x)$, and define $n = \sum_i h_i$. We can then write the estimated means and covariances as:

$$\mu_x = \frac{\sum_i h_i x_i}{n}, \ \sigma_x^2 = \frac{\sum_i h_i (x_i - \mu_x)^2}{n}, \ \sigma_{xy} = \frac{\sum_i h_i (x_i - \mu_x)(y_i - \mu_y)}{n}$$

$$\mu_y = \frac{\sum_i h_i y_i}{n}, \ \sigma_y^2 = \frac{\sum_i h_i (y_i - \mu_y)^2}{n}, \ \sigma_{y|x}^2 = \sigma_y^2 - \frac{\sigma_{xy}^2}{\sigma_x^2}.$$

We use the data covariances to express the conditional expectations and their estimated variances:

$$\hat{y} = \mu_y + \frac{\sigma_{xy}}{\sigma_x^2}(x - \mu_x), \qquad \sigma_{\hat{y}}^2 = \frac{\sigma_{y|x}^2}{n^2}\left(\sum_i h_i^2 + \frac{(x - \mu_x)^2}{\sigma_x^2}\sum_i h_i^2 \frac{(x_i - \mu_x)^2}{\sigma_x^2}\right) \qquad (10)$$

### 4.1 Setting the Smoothing Parameter $k$

There are a number of ways one can set $k$, the smoothing parameter. The method used by Cleveland et al. (1988) is to set $k$ such that the reference point being predicted has a predetermined amount of support, that is, $k$ is set so that $n$ is close to some target value. This has the disadvantage of requiring assumptions about the noise and smoothness of the function being learned. Another technique, used by Schaal and Atkeson (1994), sets $k$ to minimize the crossvalidated error on the training set. A disadvantage of this technique is that it assumes the distribution of the training set is representative of $P(x)$, which it may not be in an active learning situation. A third method, also described by Schaal and Atkeson (1994), is to set $k$ so as to minimize the estimate of $\sigma_{\hat{y}}^2$ at the reference points. As $k$ decreases, the regression becomes more global. The total weight $n$ will increase (which decreases $\sigma_{\hat{y}}^2$), but so will the conditional variance $\sigma_{y|x}^2$ (which increases $\sigma_{\hat{y}}^2$). At some value of $k$, these two quantities will balance to produce a minimum estimated variance (see Figure 3). This estimate can be computed for arbitrary reference points in the domain, and the user has the option of using either a different $k$ for each reference point or a single global $k$ that minimizes the average $\sigma_{\hat{y}}^2$ over all reference points. Empirically, we found that the variance-based method gave the best performance.

### 4.2 Active Learning with Locally Weighted Regression

As with the mixture of Gaussians, we want to select $\tilde{x}$ to minimize $\left\langle \tilde{\sigma}_{\hat{y}}^2 \right\rangle$. To do this, we must estimate the mean and variance of $P(\tilde{y}|\tilde{x})$. With locally weighted regression, these are explicit: the mean is $\hat{y}(\tilde{x})$ and the variance is $\sigma_{y|\tilde{x}}^2$. The estimate of $\left\langle \tilde{\sigma}_{\hat{y}}^2 \right\rangle$ is also explicit. Defining $\tilde{h}$ as the weight assigned to $\tilde{x}$ by the kernel we can compute these expectations exactly in closed form. For the LOESS model, the learner's expected new variance is

$$\left\langle \tilde{\sigma}_{\hat{y}}^2 \right\rangle = \frac{\left\langle \tilde{\sigma}_{y|x}^2 \right\rangle}{(n+\tilde{h})^2} \left[ \sum_i h_i^2 + \tilde{h}^2 + \frac{(x - \tilde{\mu}_x)^2}{\tilde{\sigma}_x^2} \left( \sum_i h_i^2 \frac{(x_i - \tilde{\mu}_x)^2}{\tilde{\sigma}_x^2} + \tilde{h}^2 \frac{(\tilde{x} - \tilde{\mu}_x)^2}{\tilde{\sigma}_x^2} \right) \right]. \quad (11)$$

Note that, since $\sum_i h_i^2 (x_i - \mu_x)^2 = \sum_i h_i^2 x_i^2 + \mu_x^2 \sum_i h_i^2 - 2\mu_x \sum_i h_i^2 x_i$, the new expectation of Equation 11 may be efficiently computed by caching the values of $\sum_i h_i^2 x_i^2$ and $\sum_i h_i^2 x_i$. This obviates the need to recompute the entire sum for each new candidate point. The component expectations in Equation 11 are computed as follows:

$$\left\langle \tilde{\sigma}_{y|x}^2 \right\rangle = \left\langle \tilde{\sigma}_y^2 \right\rangle - \frac{\left\langle \tilde{\sigma}_{xy}^2 \right\rangle}{\tilde{\sigma}_x^2}, \quad \left\langle \tilde{\sigma}_y^2 \right\rangle = \frac{n\sigma_y^2}{n+\tilde{h}} + \frac{n\tilde{h}\left(\sigma_{y|\tilde{x}}^2 + (\hat{y}(\tilde{x}) - \mu_y)^2\right)}{(n+\tilde{h})^2},$$

$$\tilde{\mu}_x = \frac{n\mu_x + \tilde{h}\tilde{x}}{n+\tilde{h}}, \quad \left\langle \tilde{\sigma}_{xy} \right\rangle = \frac{n\sigma_{xy}}{n+\tilde{h}} + \frac{n\tilde{h}(\tilde{x} - \mu_x)(\hat{y}(\tilde{x}) - \mu_y)}{(n+\tilde{h})^2},$$

$$\tilde{\sigma}_x^2 = \frac{n\sigma_x^2}{n+\tilde{h}} + \frac{n\tilde{h}(\tilde{x} - \mu_x)^2}{(n+\tilde{h})^2}, \quad \left\langle \tilde{\sigma}_{xy}^2 \right\rangle = \left\langle \tilde{\sigma}_{xy} \right\rangle^2 + \frac{n^2 \tilde{h}^2 \sigma_{y|\tilde{x}}^2 (\tilde{x} - \mu_x)^2}{(n+\tilde{h})^4}.$$

Just as with the mixture of Gaussians, we can use the expectation in Equation 11 to guide active learning.

## 5. Experimental Results

For an experimental testbed, we used the "Arm2D" problem described by Cohn (1994). The task is to learn the kinematics of a toy 2-degree-of-freedom robot arm (see Figure 4). The inputs are joint angles $(\Theta_1, \Theta_2)$, and the outputs are the Cartesian coordinates of the tip $(X_1, X_2)$. One of the implicit assumptions of both models described here is that the noise is Gaussian in the output dimensions. To test the robustness of the algorithm to this assumption, we ran experiments using no noise, using additive Gaussian noise in the outputs, and using additive Gaussian noise in the inputs. The results of each were comparable; we report here the results using additive Gaussian noise in the inputs. Gaussian input noise corresponds to the case where the arm effectors or joint angle sensors are noisy, and results in non-Gaussian errors in the learner's outputs. The input distribution $P(x)$ is assumed to be uniform.

We compared the performance of the variance-minimizing criterion by comparing the *learning curves* of a learner using the criterion with that of one learning from random samples. The learning curves plot the mean squared error and variance of the learner as its training set size increases. The curves are created by starting with an initial sample, measuring the learner's mean squared error or estimated variance on a set of "reference" points (independent of the training set), selecting and adding a new example to the training set, retraining the learner on the augmented set, and repeating.

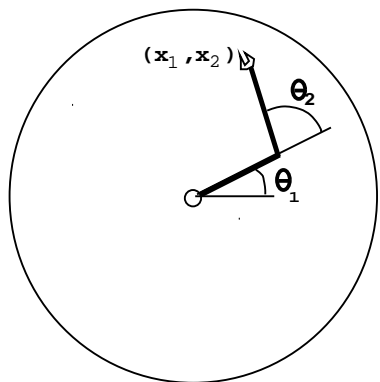

Figure 4: The arm kinematics problem. The learner attempts to predict tip position given a set of joint angles $(\theta_1, \theta_2)$.

On each step, the variance-minimizing learner chose a set of 64 unlabeled reference points drawn from input distribution $P(x)$. It then selected a query $\tilde{x} = (\theta_1, \theta_2)$ that it estimated would minimize $\left\langle \tilde{\sigma}_{y|x}^2 \right\rangle$ over the reference set. In the experiments reported here, the best $\tilde{x}$ was selected from another set of 64 "candidate" points drawn at random on each iteration.[2]

## 5.1 Experiments with Mixtures of Gaussians

With the mixtures of Gaussians model, there are three design parameters that must be considered — the number of Gaussians, their initial placement, and the number of iterations of the EM algorithm. We set these parameters by optimizing them on the learner using random examples, then used the same settings on the learner using the variance-minimization criterion. Parameters were set as follows: Models with fewer Gaussians have the obvious advantage of requiring less storage space and computation. Intuitively, a small model should also have the advantage of avoiding overfitting, which is thought to occur in systems with extraneous parameters. Empirically, as we increased the number of Gaussians, generalization improved monotonically with diminishing returns (for a fixed training set size and number of EM iterations). The test error of the larger models generally matched that of the smaller models on small training sets (where overfitting would be a concern), and continued to decrease on large training sets where the smaller networks "bottomed out." We therefore preferred the larger mixtures, and report here our results with mixtures of 60 Gaussians. We selected initial placement of the Gaussians randomly, chosen uniformly from the smallest hypercube containing all current training examples. We arbitrarily chose the identity matrix as an initial covariance matrix. The learner was surprisingly sensitive to the number of EM iterations. We examined a range of 5 to 40 iterations of the EM algorithm per step. Small numbers of iterations (5-10) appear insufficient to allow convergence with large training sets, while large numbers of iterations (30-40) degraded performance on small training sets. An ideal training regime would employ some form of regularization, or would examine the degree of change between iterations to detect convergence; in our experiments, however, we settled on a fixed regime of 20 iterations per step.

---

2. As described earlier, we could also have selected queries by hillclimbing on $\partial \left\langle \tilde{\sigma}_{y|x}^2 \right\rangle / \partial \tilde{x}$; in this low dimensional problem it was more computationally efficient to consider a random candidate set.

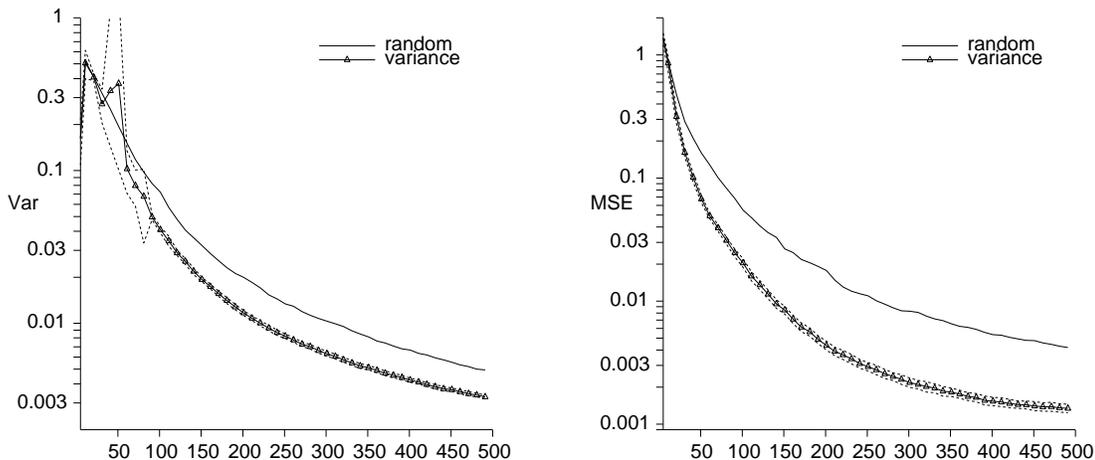

Figure 5: Variance and MSE learning curves for mixture of 60 Gaussians trained on the Arm2D domain. Dotted lines denote standard error for average of 10 runs, each started with one initial random example.

Figure 5 plots the variance and MSE learning curves for a mixture of 60 Gaussians trained on the Arm2D domain with 1% input noise added. The estimated model variance using the variance-minimizing criterion is significantly better than that of the learner selecting data at random. The mean squared error, however, exhibits even greater improvement, with an error that is consistently 1/3 that of the randomly sampling learner.

## 5.2 Experiments with LOESS Regression

With LOESS, the design parameters are the the size and shape of the kernel. As described earlier, we arbitrarily chose to work with a Gaussian kernel; we used the variance-based method for automatically selecting the kernel size.

In the case of LOESS, both the variance and the MSE of the learner using the variance-minimizing criterion are significantly lower than those of the learner selecting data randomly. It is worth noting that on the Arm2D domain, this form of locally weighted regression also significantly outperforms both the mixture of Gaussians and the neural networks discussed by Cohn (1994).

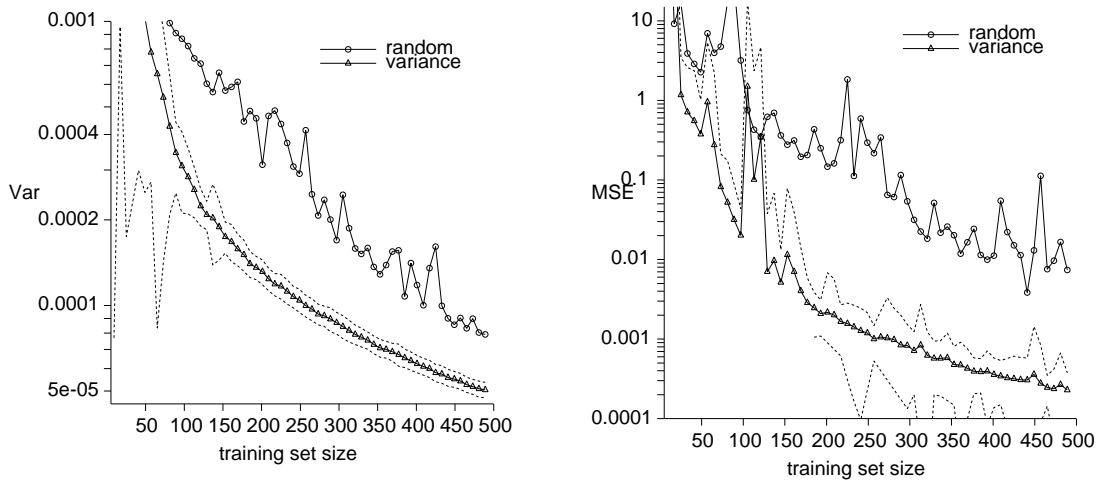

Figure 6: Variance and MSE learning curves for LOESS model trained on the Arm2D domain. Dotted lines denote standard error for average of 60 runs, each started with a single initial random example.

### 5.3 Computation Time

One obvious concern about the criterion described here is its computational cost. In situations where obtaining new examples may take days and cost thousands of dollars, it is clearly wise to expend computation to ensure that those examples are as useful as possible. In other situations, however, new data may be relatively inexpensive, so the computational cost of finding optimal examples must be considered.

Table 1 summarizes the computation times for the two learning algorithms discussed in this paper.[3] Note that, with the mixture of Gaussians, training time depends linearly on the number of examples, but prediction time is independent. Conversely, with locally weighted regression, there is no "training time" per se, but the cost of additional examples accrues when predictions are made using the training set.

While the training time incurred by the mixture of Gaussians may make it infeasible for selecting optimal action learning actions in realtime control, it is certainly fast enough to be used in many applications. Optimized, parallel implementations will also enhance its utility.[4] Locally weighted regression is certainly fast enough for many control applications, and may be made faster still by optimized, parallel implementations. It is worth noting that, since the prediction speed of these learners depends on their training set size, optimal data selection is doubly important, as it creates a parsimonious training set that allows faster predictions on future points.

---

3. The times reported are "per reference point" and "per candidate per reference point"; overall time must be computed from the number of candidates and reference points examined. In the case of the LOESS model, for example, with 100 training points, 64 reference points and 64 candidate points, the time required to select an action would be $(58 + 0.16 \times 100) \times 4096 \mu$seconds, or about 0.3 seconds.
4. It is worth mentioning that approximately half of the training time for the mixture of Gaussians is spent computing the correction factor in Equation 8. Without the correction, the learner still computes $P(y|x)$, but does so by modeling the training set distribution rather than the reference distribution. We have found however, that for the problems examined, the performance of such "uncorrected" learners does not differ appreciably from that of the "corrected" learners.

|  | Training | Evaluating Reference | Evaluating Candidates |
|---|---|---|---|
| Mixture | $3.9 + 0.05m$ sec | 15000 $\mu$sec | 1300 $\mu$sec |
| LOESS | - | $92 + 9.7m$ $\mu$sec | $58 + 0.16m$ $\mu$sec |

Table 1: Computation times on a Sparc 10 as a function of training set size $m$. Mixture model had 60 Gaussians trained for 20 iterations. Reference times are per reference point; candidate times are per candidate point per reference point.

## 6. Discussion

Mixtures of Gaussians and locally weighted regression are two statistical models that offer elegant representations and efficient learning algorithms. In this paper we have shown that they also offer the opportunity to perform active learning in an efficient and statistically correct manner. The criteria derived here can be computed cheaply and, for problems tested, demonstrate good predictive power. In industrial settings, where gathering a single data point may take days and cost thousands of dollars, the techniques described here have the potential for enormous savings.

In this paper, we have only considered function approximation problems. Problems requiring classification could be handled analogously with the appropriate models. For learning classification with a mixture model, one would select examples so as to maximize discriminability between Gaussians; for locally weighted regression, one would use a logistic regression instead of the linear one considered here (Weisberg, 1985).

Our future work will proceed in several directions. The most important is active bias minimization. As noted in Section 2, the learner's error is composed of both bias and variance. The variance-minimizing strategy examined here ignores the bias component, which can lead to significant errors when the learner's bias is non-negligible. Work in progress examines effective ways of measuring and optimally eliminating bias (Cohn, 1995); future work will examine how to jointly minimize both bias and variance to produce a criterion that truly minimizes the learner's expected error.

Another direction for future research is the derivation of variance- (and bias-) minimizing techniques for other statistical learning models. Of particular interest is the class of models known as "belief networks" or "Bayesian networks" (Pearl, 1988; Heckerman, Geiger, & Chickering, 1994). These models have the advantage of allowing inclusion of domain knowledge and prior constraints while still adhering to a statistically sound framework. Current research in belief networks focuses on algorithms for efficient inference and learning; it would be an important step to derive the proper criteria for learning *actively* with these models.

## Appendix A. Notation

| | General |
|---|---|
| $X$ | input space |
| $Y$ | output space |
| $x$ | an arbitrary point in the input space |
| $y$ | true output value corresponding to input $x$ |
| $\hat{y}$ | predicted output value corresponding to input $x$ |
| $x_i$ | "input" part of example $i$ |
| $y_i$ | "output" part of example $i$ |
| $m$ | the number of examples in the training set |
| $\tilde{x}$ | specified input of a query |
| $\tilde{y}$ | the (possibly not yet known) output of query $\tilde{x}$ |
| $\sigma_{\hat{y}}^2$ | estimated variance of $\hat{y}$ |
| $\tilde{\sigma}_{\hat{y}}^2$ | new variance of $\hat{y}$, after example $(\tilde{x}, \tilde{y})$ has been added |
| $\left\langle \tilde{\sigma}_{\hat{y}}^2 \right\rangle$ | the expected value of $\tilde{\sigma}_{\hat{y}}^2$ |
| $P(x)$ | the (known) natural distribution over $x$ |

| | Neural Network |
|---|---|
| $w$ | a weight vector for a neural network |
| $\hat{w}$ | estimated "best" $w$ given a training set |
| $f_{\hat{w}}()$ | function computed by neural network given $\hat{w}$ |
| $S^2$ | average estimated noise in data, used as an estimate for $\sigma_y^2$ |

| | Mixture of Gaussians |
|---|---|
| $N$ | total number of Gaussians |
| $g_i$ | Gaussian number $i$ |
| $n_i$ | total point weighting attributed to Gaussian $i$ |
| $\mu_{x,i}$ | estimated $x$ mean of Gaussian $i$ |
| $\mu_{y,i}$ | estimated $y$ mean of Gaussian $i$ |
| $\sigma_{x,i}^2$ | estimated $x$ variance of Gaussian $i$ |
| $\sigma_{y,i}^2$ | estimated $y$ variance of Gaussian $i$ |
| $\sigma_{xy,i}$ | estimated $xy$ covariance of Gaussian $i$ |
| $\sigma_{y|x,i}^2$ | estimated $y$ variance of Gaussian $i$, given $x$ |
| $P(x,y|i)$ | joint distribution of input-output pair given Gaussian $i$ |
| $P(x|i)$ | distribution $x$ being given Gaussian $i$ |
| $h_i$ | weight of a given point that is attributed to Gaussian $i$ |
| $\tilde{h}_i$ | weight of new point $(\tilde{x}, \tilde{y})$ that is attributed to Gaussian $i$ |

| | Locally Weighted Regression |
|---|---|
| $k$ | kernel smoothing parameter |
| $h_i$ | weight given to example $i$ by kernel centered at $x$ |
| $n$ | sum of weights given to all points by kernel |
| $\mu_x$ | mean of inputs, weighted by kernel centered at $x$ |
| $\mu_y$ | mean of outputs, weighted by kernel centered at $x$ |
| $\tilde{h}$ | weight of new point $(\tilde{x}, \tilde{y})$ given kernel centered at $x$ |

## Acknowledgements

David Cohn's current address is: Harlequin, Inc., One Cambridge Center, Cambridge, MA 02142 USA. Zoubin Ghahramani's current address is: Department of Computer Science, University of Toronto, Toronto, Ontario M5S 1A4 CANADA. This work was funded by NSF grant CDA-9309300, the McDonnell-Pew Foundation, ATR Human Information Processing Laboratories and Siemens Corporate Research. We are deeply indebted to Michael Titterington and Jim Kay, whose careful attention and continued kind help allowed us to make several corrections to an earlier version of this paper.